\documentclass{article}

    \PassOptionsToPackage{numbers, sort&compress}{natbib}

    \usepackage[preprint]{neurips_2023}

\usepackage[utf8]{inputenc} 
\usepackage[T1]{fontenc}    
\usepackage{hyperref}       
\usepackage{url}            
\usepackage{booktabs}       
\usepackage{amsfonts}       
\usepackage{nicefrac}       
\usepackage{microtype}      
\usepackage{xcolor}         

\usepackage{graphicx}
\usepackage{multirow}
\usepackage[title]{appendix}
\usepackage[aboveskip=-2pt,belowskip=-10pt]{caption}
\usepackage{subcaption}

\title{On Calibration of Modern Quantized Efficient Neural Networks}

\author{%
  Joey Kuang \\
  Vision and Image Processing Research Group \\
  University of Waterloo \\
  Waterloo, Canada \\
  \texttt{jjykuang@uwaterloo.ca} \\
  \And
  Alexander Wong \\
  Vision and Image Processing Research Group \\
  Waterloo Artificial Intelligence Institute \\
  DarwinAI Corp. \\ 
  University of Waterloo \\
  Waterloo, Canada \\
  \texttt{a28wong@uwaterloo.ca} \\
}

\begin{document}

\maketitle

\begin{abstract}
  We explore calibration properties at various precisions for three architectures: ShuffleNetv2, GhostNet-VGG, and MobileOne; and two datasets: CIFAR-100 and PathMNIST. The quality of calibration is observed to track the quantization quality; it is well-documented that performance worsens with lower precision, and we observe a similar correlation with poorer calibration. This becomes especially egregious at 4-bit activation regime. GhostNet-VGG is shown to be the most robust to overall performance drop at lower precision. We find that temperature scaling can improve calibration error for quantized networks, with some caveats. We hope that these preliminary insights can lead to more opportunities for explainable and reliable EdgeML.
\end{abstract}

\section{Introduction}\label{section:intro}
Enabling critical decision-making machine learning applications on the edge, such as point-of-care testing, increases efficiency and accessibility \cite{khan_artificial_2023}. Deploying models may require quantization, which calls for consideration to performance drop. Meanwhile, critical decision-making leaves little room for predictive error, thus emphasizing the role of model uncertainty. 

The seminal work of Guo \etal \cite{guo_calibration_2017} revealing the calibration dilemma present in modern neural networks brought forth calibration estimators and methods, and studies of intrinsic calibration properties of an architecture \cite{laves_well-calibrated_2019,minderer_revisiting_2021,muller_when_2020,ovadia_can_2019,perez-lebel_beyond_2023,wang_rethinking_2021}.
There are few works that intersect the model's quantization behaviour and calibration quality. Studies regarding model quantization have focused on reducing undesirable quantization noise by way of novel quantization methods \cite{dong_hawq_2019,defossez_differentiable_2022,vedaldi_hmq_2020,liu_sharpness-aware_2023,park_nipq_2022,zhou_incremental_2017}. Intuitively, we recognize that error in quantized outputs can cause decision swaps and change in confidence. In turn, we suspect possible consequences on calibration properties. One work by Xia \etal \cite{xia_underexplored_2021} observes that better calibrated models tend to provide worse post-quantization performance. In this extended abstract, we continue to explore the relationship between quantization and calibration: what do model calibration properties look like across various precision bitwidths and how well does a quantized model receive a post-hoc calibration method?

\begin{figure}[t]
   \begin{center}
      \includegraphics[width=0.9\linewidth]{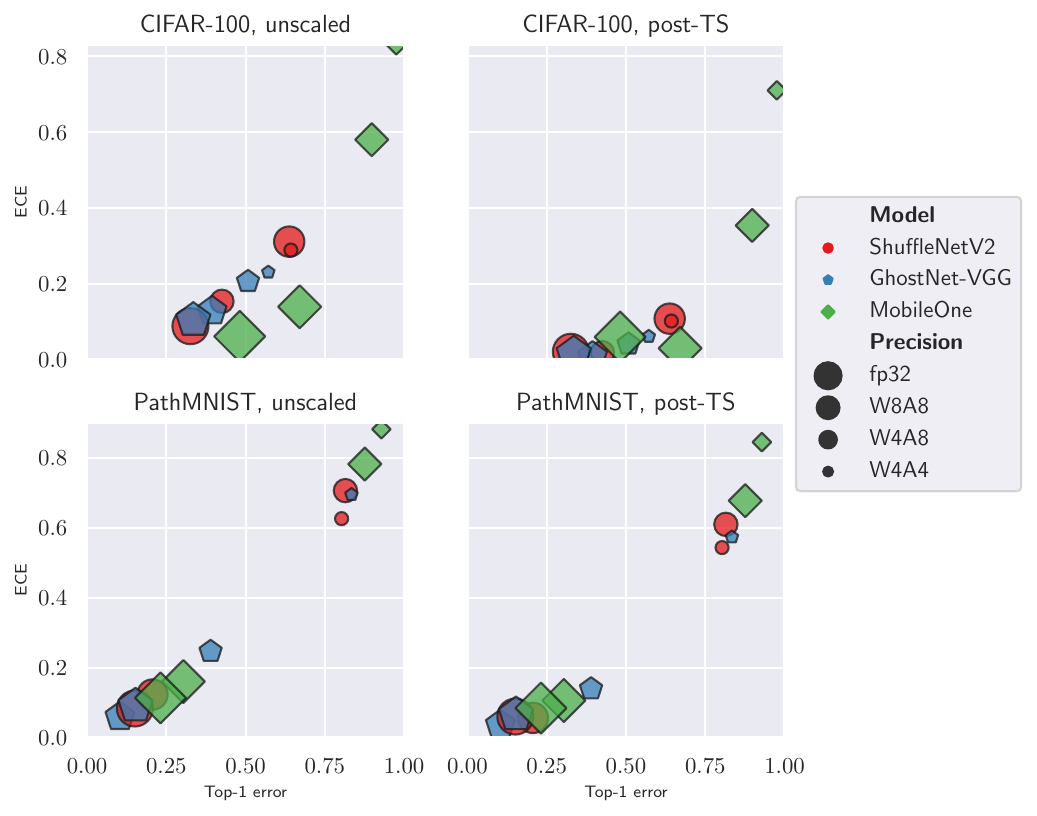}
   \end{center}
   \caption{ECE and top-1 error before (left column) and after (right column) temperature scaling. Model backbones are grouped by color and precision levels are grouped by marker size. Lower bitwidth models are correlated with both poorer accuracy and poorer calibration.}
   \label{fig:pareto}
\end{figure}

\section{Experiments}\label{section:expts}

We conduct our analyses on the CIFAR-100 (32x32) dataset and PathMNIST2D (28x28) \cite{krizhevsky_learning_nodate,yang_medmnist_2023}. Using biomedical image data provides context for understanding calibration properties due to related high-stakes tasks, such as lesion detection \cite{lara2023unraveling}. PathMNIST2D (9 classes, 107k samples total) was sourced from research on predicting survival from colorectal cancer histology slides.

To investigate from a more practical perspective, we select architectures designed for mobile targets and likely to face quantization. Due to the small image sizes involved, we make modifications to the model based on the source paper or other relevant works. For ShuffleNetv2, We follow AugShuffleNet \cite{ye_augshufflenet_2022} which was modified for CIFAR datasets. While the main GhostNet is built on MobileNetV3 \cite{howard_searching_2019}, Han \etal \cite{han_ghostnet_2020} run toy experiments on CIFAR datasets with a VGG16-like network provided by \cite{noauthor_torch_nodate}. With best judgement, we reduce depth, stride, and width of the MobileOne S0 variant to avoid excessive downsampling and overcomplexity \cite{vasu_improved_2022}.

\subsection{Training}
We train models for 200 epochs with an effective batch size of 32 using SGD with momentum ($\beta=0.9$) \cite{sutskever_importance_2013}. 
We initialize with He initialization \cite{he_delving_2015} and train on cross-entropy loss with a weight decay set to $10^{-4}$. Note that we specifically avoid other training time regularization methods to avoid confounding with post-hoc calibration \cite{guo_calibration_2017}.
We initialize our learning rate at 0.1 and decay it using a cosine schedule with warm restarts; we follow Scenario 6 of CIFAR-10 settings from \cite{loshchilov_sgdr_2017}. We apply random crop and random horizontal flip. 

We follow a modified implementation of the Pytorch post-training quantization library by Xia \etal \cite{xia_underexplored_2021} to quantize our weights symmetrically per-channel with a moving average min-max observer; activations are quantized asymmetrically per-tensor with a histogram-based observer.
We perform quantization at 3 configurations: W8/A8, W4/A8, W4/A4. Each configuration is quantized at 3 trials with randomly sampled training batches for calibrating ranges, where total calibration size is 1024 \cite{nagel_white_2021}. 
All quantized \textit{inference} is performed at 8-bit. We choose temperature scaling (TS) as our sole post-hoc calibration method \cite{guo_calibration_2017} and Expected Calibration Error (ECE) as the primary metric for calibration quality (lower is better) \cite{naeini_calibrated_2015}.

\begin{figure}[!ht]
   \begin{center}
      \includegraphics[width=0.8\linewidth]{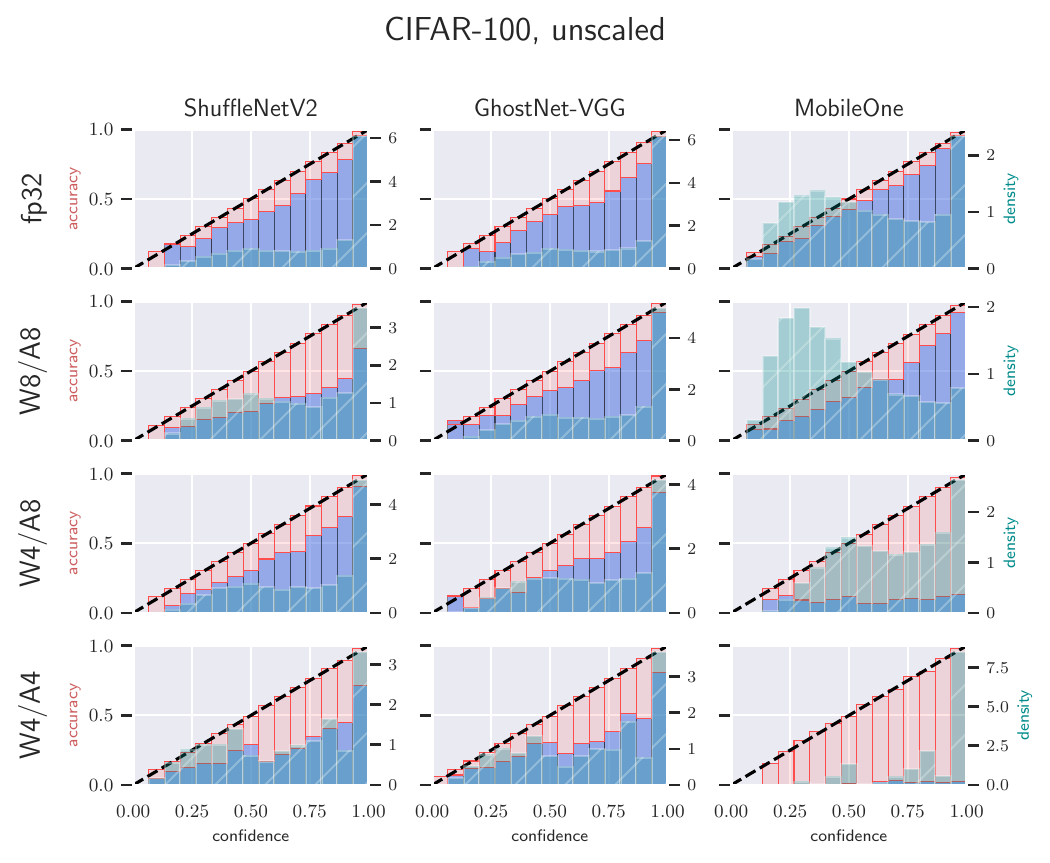}
   \end{center}
   \caption{Reliability grid (blue and red) and confidence distribution (striped teal) of CIFAR-100 models (by column) of each precision level (by row) before temperature scaling. Larger red bars indicate greater (worse) confidence-accuracy gap. Although MobileOne FP32 is well-calibrated, more so than GhostNet-VGG FP32 and ShuffleNetV2 FP32, its quantized counterparts are very poorly calibrated.}
   \label{fig:cifar100_NS_reliability_grid}
\end{figure}

\begin{figure}[!ht]
   \begin{center}
      \includegraphics[width=0.8\linewidth]{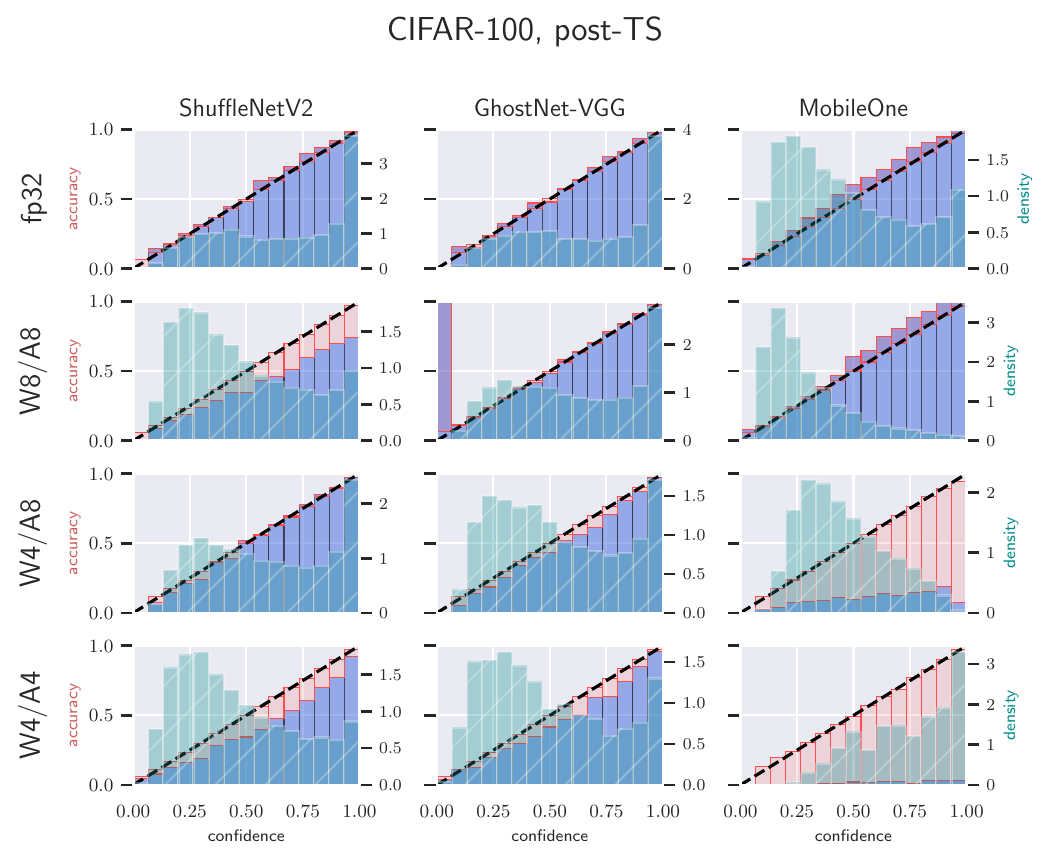}
   \end{center}
   \caption{Reliability grid (blue and red) and confidence distribution (striped teal) of CIFAR-100 models (by column) of each precision level (by row) after temperature scaling. TS does well to correct confidence-accuracy gaps of ShuffleNetV2 and GhostNet-VGG, while extremely poorly calibrated pre-TS 4-bit weight MobileOne models cannot recover.}
   \label{fig:cifar100_TS_reliability_grid}
\end{figure}

\begin{figure}[!ht]
   \begin{center}
      \includegraphics[width=0.8\linewidth]{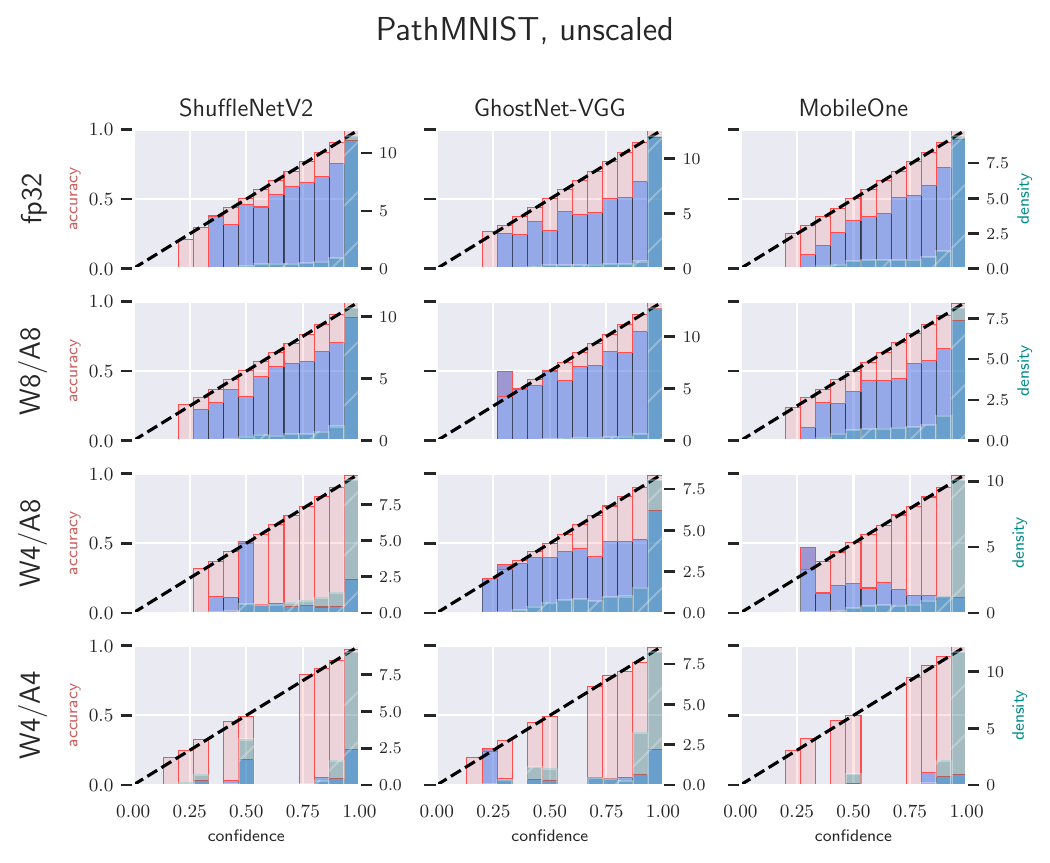}
   \end{center}
   \caption{Reliability grid (blue and red) and confidence distribution (striped teal) of PathMNIST models (by column) of each precision level (by row) before temperature scaling. 4-bit weight model performance and calibration consistently deteriorates for models using inverted residual blocks (ShuffleNetV2 and MobileOne).}
   \label{fig:pathmnist_NS_reliability_grid}
\end{figure}

\begin{figure}[!ht]
   \begin{center}
      \includegraphics[width=0.8\linewidth]{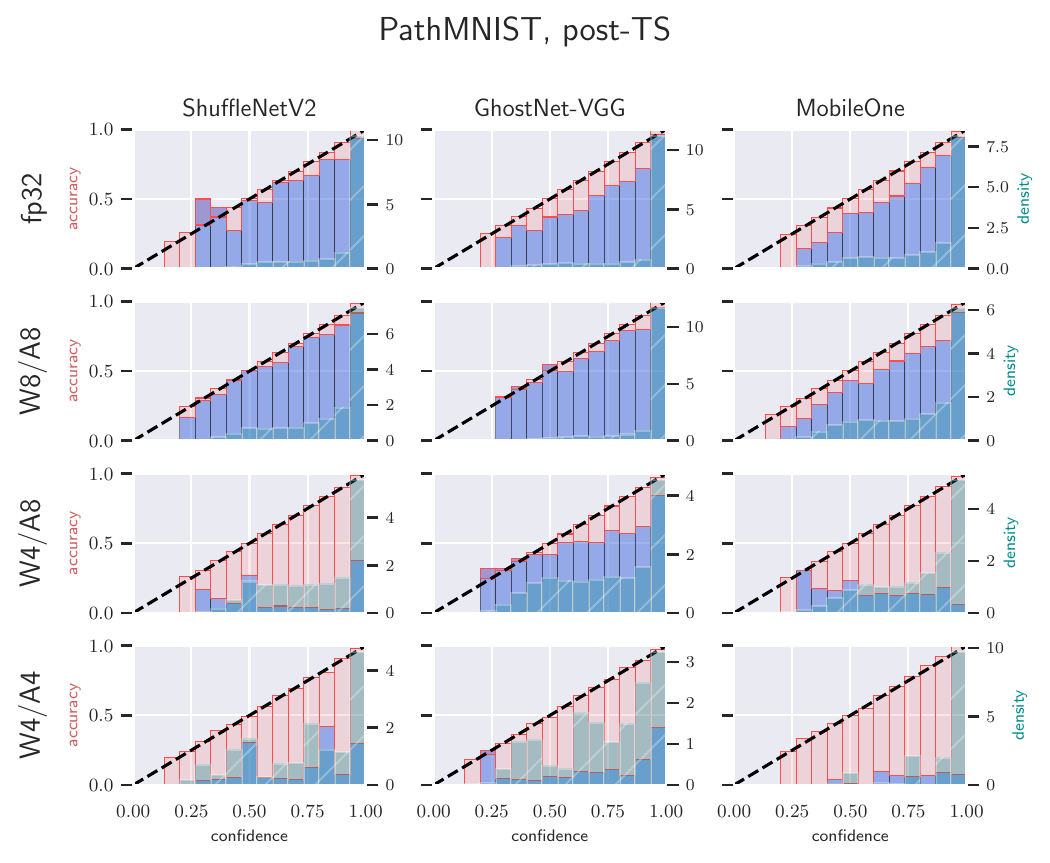}
   \end{center}
   \caption{Reliability grid (blue and red) and confidence distribution (striped teal) of PathMNIST models (by column) of each precision level (by row) before temperature scaling. While confidence-accuracy gap tends to diminish post-TS, improvement is not as noticeable under PathMNIST settings, especially when the initial gap is incredibly large.}
   \label{fig:pathmnist_TS_reliability_grid}
\end{figure}

\section{Results and Discussion}\label{section:results}

Figure \ref{fig:pareto} shows that calibration properties are consistent between models and their precision variants across datasets. Minderer \etal \cite{minderer_revisiting_2021} has suggested that architecture is a key determinant of calibration properties. The exact factors are unclear, though they do find that it is not related to model size nor amount of pretraining. We observe that GhostNet-VGG is the most robust to both quantization-induced accuracy decrease and ECE increase, while lower-bit ShuffleNetV2 and MobileOne model performance suffers. Notably, these two backbones consist of inverted residual blocks employing depthwise-separable convolutions, which have been shown to suffer from larger accumulation of quantization errors \cite{yun_all_2021}.

We find that TS is at some point limited by how well the model performs in the first place. First, we observe expected TS performance between Figure \ref{fig:cifar100_NS_reliability_grid} and Figure \ref{fig:cifar100_TS_reliability_grid} for most models, where post-TS models are well-calibrated. However, TS is unable to correct the 4-bit MobileOne models that have a larger initial confidence-accuracy gap. This is further shown by Figure \ref{fig:pathmnist_NS_reliability_grid}, emphasized on PathMNIST: higher-precision models are accurate, well-calibrated (vs. counterpart trained on CIFAR-100), and refined (making more confident decisions) resulting in a tight cluster near the origin in Figure \ref{fig:pareto}. Despite this success, all W4/A4 models perform terribly w.r.t. ECE and Top-1 and large confidence-accuracy gaps remain in post-TS models in Figures \ref{fig:cifar100_TS_reliability_grid} and \ref{fig:pathmnist_TS_reliability_grid}. This is consistent with \cite{xia_underexplored_2021}; in addition to well-calibrated floating point models posting worse post-quantization performance, their calibration quality appears to drop as well. Their conclusions were drawn from exploring ResNet models, which points to residual connections as a suspect for both of our findings. Interestingly, while prior work \cite{guo_calibration_2017,minderer_revisiting_2021} has observed that within a model family (e.g., grouped by size), a model with lower classification error tends to have higher calibration error, we show the exact opposite within a set of model's quantized counterparts. We also find the large discrepancy in quantization performance of ShuffleNetV2 particularly interesting between PathMNIST and CIFAR-100 settings; it may be worth investigating in relation to ECE biases and dataset attributes. 

Further analysis includes investigating the correlation between the pre- and post-TS ECE delta and the overall quantization quality to understand possible parallelisms between calibration properties and quantization behaviour. While we provided an initial discussion on certain design patterns contributing to a deployable model's dismay, additional exploration on intrinsic architecture properties relating to quantization \textit{and} calibration, training policies, quantization methods, and calibrators, are paramount to outline best practices for \textit{trustworthy} EdgeML solutions.

{\small
\bibliographystyle{unsrtnat}
\bibliography{egbib}
}


\end{document}